\documentclass[runningheads]{llncs}

\usepackage{eccv}

\usepackage{eccvabbrv}

\usepackage{graphicx}
\usepackage{booktabs}
\usepackage{graphicx}
\usepackage{amsmath}
\usepackage{booktabs}

\usepackage[noend]{algpseudocode}
\usepackage{algorithmicx,algorithm}
\usepackage{colortbl}
\usepackage{multirow}
\usepackage{sidecap}
\usepackage{wrapfig}
\usepackage{bbding}

\usepackage[accsupp]{axessibility}  

\usepackage{hyperref}

\usepackage{orcidlink}

\begin{document}

\title{Pathology-knowledge Enhanced \\ Multi-instance Prompt Learning for \\ Few-shot Whole Slide Image Classification} 

\titlerunning{Pathology-knowledge Enhanced Multi-instance Prompt Learning}

\author{Linhao~Qu\inst{1}\orcidlink{0000-0001-8815-7050} \and
Dingkang~Yang\inst{2}\orcidlink{0000-0003-1829-5671} \and
Dan~Huang\inst{3}\and
Qinhao~Guo\inst{4}\and
Rongkui~Luo\inst{5}\orcidlink{0000-0002-4118-7225}\and
Shaoting~Zhang\inst{1}\and
Xiaosong~Wang\inst{1}\textsuperscript{\Envelope}\orcidlink{0000-0002-3840-5658}}

\authorrunning{L. Qu et al.}

\institute{Shanghai Artificial Intelligence Laboratory, Shanghai, China \and
Academy for Engineering and Technology, Fudan University, Shanghai, China \and
Department of Pathology, Fudan University Shanghai Cancer Center, China \and
Department of Gynecologic Oncology, Fudan University Shanghai Cancer Center \and
Department of Pathology, Zhongshan Hospital, Fudan University, Shanghai, China  \\
\email{wangxiaosong@pjlab.org.cn};
}

\maketitle

\begin{abstract}
Current multi-instance learning algorithms for pathology image analysis often require a substantial number of Whole Slide Images for effective training but exhibit suboptimal performance in scenarios with limited learning data. In clinical settings, restricted access to pathology slides is inevitable due to patient privacy concerns and the prevalence of rare or emerging diseases. The emergence of the Few-shot Weakly Supervised WSI Classification accommodates the significant challenge of the limited slide data and sparse slide-level labels for diagnosis. Prompt learning based on the pre-trained models (\eg, CLIP) appears to be a promising scheme for this setting; however, current research in this area is limited, and existing algorithms often focus solely on patch-level prompts or confine themselves to language prompts. This paper proposes a multi-instance prompt learning framework enhanced with pathology knowledge, \ie, integrating visual and textual prior knowledge into prompts at both patch and slide levels. The training process employs a combination of static and learnable prompts, effectively guiding the activation of pre-trained models and further facilitating the diagnosis of key pathology patterns. Lightweight Messenger (self-attention) and Summary (attention-pooling) layers are introduced to model relationships between patches and slides within the same patient data. Additionally, alignment-wise contrastive losses ensure the feature-level alignment between visual and textual learnable prompts for both patches and slides. Our method demonstrates superior performance in three challenging clinical tasks, significantly outperforming comparative few-shot methods.
  \keywords{Pathology image analysis \and Prompt learning}
\end{abstract}

\section{Introduction}
\label{sec:intro}
Deep learning for pathology Whole Slide Image (WSI) classification offers a rapid, objective, and precise diagnostic approach, enhancing prognosis prediction and treatment response assessment \cite{24,92,93,94,95,97}. Given the gigapixel resolution of WSIs, segmentation into multiple patches is essential for processing. However, obtaining detailed annotations for each patch is challenging. Instead, it often results in slide-level ground truth labels. Multiple Instance Learning (MIL), a prominent weakly supervised paradigm, has emerged as a key technique for WSI classification \cite{26,27,34,37}. In MIL, each WSI is treated as a bag, with patches as instances. During training, only slide-level labels are available, necessitating a comprehensive model to capture relationships between patches and slide labels for accurate classification \cite{3}.
\begin{figure}[t!]
  \centering
   \includegraphics[width=0.6\linewidth]{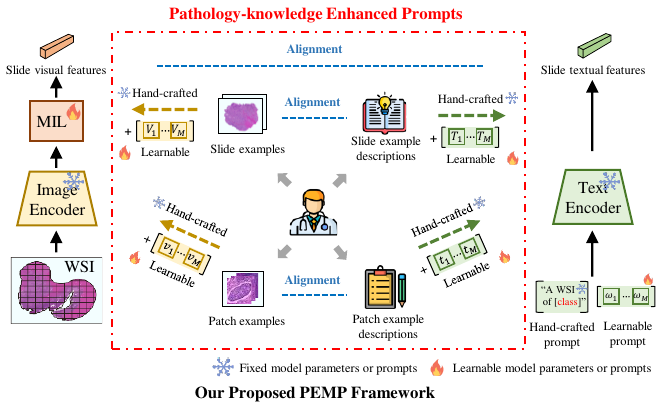}
   \caption{Existing methods such as CoOp~\cite{56} do not fully consider task-related pathology visual features and its association with specific terms in FSWC. We utilize aligned task-specific image examples and language descriptions to enhance visual and textual prompt learning at both patch and slide levels.}
   \label{figure1}
\end{figure}


MIL algorithms achieve slide-level classification by aggregating patch-level features or predictions. Though effective, existing algorithms typically demand a considerable number of slides for practical training, performing inadequately in scenarios with limited samples. In clinical practice, challenges such as preserving patient privacy, the rarity of certain diseases, and emerging conditions, particularly for prognosis and rare tumor subtypes, hinder the acquisition of a substantial quantity of WSIs \cite{24,96,27}. Recently, the challenging problem of Few-shot weakly Supervised WSI Classification (FSWC) has emerged \cite{83}. FSWC is characterized by the limited availability of training WSIs (\eg, typically 2, 4, 8, 16, or 32 per class) and sparse slide-level annotations, with unknown fine-grained annotations for numerous patches in each WSI.

In image classification, recent advancements highlight the effectiveness of Vision-Language Models (VLMs) like CLIP \cite{53}, BLIP \cite{84}, and Flamingo \cite{85}. These models undergo training on large-scale language-image pairs using alignment contrastive learning, demonstrating notable zero-shot and few-shot transfer learning capabilities. In the pathology image domain, despite valuable contributions from initiatives like MI-Zero \cite{66} and PLIP \cite{67} in constructing VLMs for computational pathology, challenges including ethical concerns, annotation scarcity, and storage limitations make it impractical to train foundational models for each specific downstream task \cite{83}. Furthermore, existing VLMs exhibit restricted direct zero-shot transfer capabilities for specific downstream tasks \cite{86,87,88}. Considering these challenges, the application of existing VLMs for developing prompt learning algorithms tailored to few-shot downstream tasks appears promising. Prompt learning involves adapting a pre-trained and then fixed VLM to new tasks by integrating task-specific prompts as additional model inputs. In the context of few-shot learning, these prompts can be either fixed or trainable, with trainable prompts enabling contrastive learning between visual and language features \cite{56,89}.

Presently, research on prompt learning for FSWC is limited \cite{83}. Previous efforts, such as MI-Zero \cite{66}, PLIP \cite{67}, and CITE \cite{90}, have predominantly focused on the utilization of ``patch-level'' prompts—textual prompts associated with pathology regions of interest. Nonetheless, \textit{visual or textual prompts in domain-specific downstream tasks may not be understandable (often unseen) for the pre-trained model and thus may not properly activate the relevant features of the pre-trained model.}
The TOP framework \cite{83} represents the initial endeavor to implement prompt learning by introducing language-based prior descriptions in the prompts. 
However, pathology images encompass highly specialized visual features and corresponding pathology-specific terms, posing a challenge for networks to comprehensively understand and learn pathology-related features and specialized descriptions solely through language explanations for classification.

To address this issue, we propose \textbf{P}athology-knowledge \textbf{E}nhanced \textbf{M}ulti-instance \textbf{P}rompt Learning for FSWC named \textbf{PEMP}, with a concise architecture depicted in Fig.~\ref{figure1}. PEMP is a multi-instance prompt learning framework specifically designed for the FSWC problem, incorporating both visual and textual samples (as a form of prior knowledge) at both patch and slide levels. Drawing inspiration from pathologists learning pathology images and language knowledge from textbooks, we introduce task-specific image examples and corresponding language descriptions, injected in both the visual and textual prompts. 
During prompt learning, the model is fed with a more understandable and effective guide (in image examples and plain words) and then activates more meaningful features for the diagnosis of key patch-level and slide-level pathology patterns in scenarios with limited training samples. 

The contribution of this work is threefold:
\begin{itemize}
\item We propose a multi-instance prompt learning framework, PEMP, introducing vision and text \textbf{prior knowledge in the designed prompts} at both patch and slide levels. A combination of static and learnable prompts is employed in the training, effectively adapting the pre-trained model to focus on key pathology patterns in patches and slides.

\item We explore the \textbf{alignment of visual and textual prompts} using contrastive losses at both patch and slide levels. We design lightweight Messenger (self-attention) and Summary (attention-pooling) layers to model relationships of patch-to-patch and patch-to-slide within the same patient data. 

\item We evaluate PEMP's performance in addressing three challenging pathology tasks in clinical practice, 
including predicting patient survival and lymph node metastasis in early cervical cancer H\&E slides and subtype classification for a rare tumor (round cell tumor). 
PEMP achieves superior performance in all these tasks under the few-shot setting, notably outperforming all comparative methods by a large margin (4\% on average). 

\end{itemize}

\section{Related Work}
\subsection{MIL for WSI Classification}
Current MIL algorithms for WSI classification can be broadly categorized as instance-based \cite{1,2,3,29,50} and bag-based \cite{8,9,10,11,12,13,14,15,16,30,37,42,51,52} methods. Instance-based methods involve generating pseudo-labels for each patch and training patch classifiers to score all patches within a slide, whereas bag-based methods, the current mainstream, extract patch features and employ techniques like attention weighting \cite{8,9,10,11,12,13,14,37} to aggregate these features into slide-level features. While existing algorithms perform well with ample training data, they lack specificity for FSWC tasks, struggling with limited samples. Our proposed PEMP, falling under the bag-based MIL paradigm, distinguishes itself by innovatively addressing FSWC through VLMs and prompt learning, deviating from traditional bag-based MIL approaches.

\subsection{VLM-Based Prompt Learning}
VLMs, comprising image and text encoders, are trained on large-scale image-text pairs using alignment contrastive learning to establish a shared feature space. Pre-trained VLMs like CLIP \cite{53}, BLIP \cite{84}, and Flamingo \cite{85} exhibit robust zero-shot and few-shot transferability in image recognition tasks. However, limited accuracy in zero-shot transfer has fueled interest in few-shot learning based on VLMs, where the prompt learning paradigm, demonstrating strong performance \cite{86,87,88}, is emerging. Prompt learning leverages a pre-trained VLM in the adaptation to new downstream tasks by incorporating task-specific prompts through enriched model inputs. These prompts could be static with manual designs or learnable vectors\cite{56,89}. 

In pathology, recent research emphasizes constructing specialized VLMs, such as MI-Zero \cite{66}, PLIP \cite{67}, and QUILT\cite{100}. However, privacy concerns, annotation scarcity, and storage demands make training separate foundational models for each downstream task impractical. Existing VLMs exhibit limited zero-shot transferability, especially for rare diseases. Unlike fine-tuning VLMs at the data level or constructing large datasets like these works, we introduce a novel prompt learning paradigm based on VLMs (e.g., CLIP with frozen parameters) to address the FSWC problem. Importantly, our scheme is versatile across various foundation models.
Moreover, MI-Zero, PLIP and QUILT mainly contribute by collecting large datasets to construct pathology VLMs but lack effective methods for FSWC. Their simple adapting strategies for FSWC (e.g., LinearProbe and Attention Pooling) limit their performance in challenging clinical tasks.

In pathology, prompt learning research for FSWC is limited. Prior works like MI-Zero \cite{66}, PLIP \cite{67}, and CITE \cite{90} only explore ``patch-level'' prompts, limiting slide-level classification performance. TOP \cite{83} introduces prompt learning at both patch and slide levels but lacks task-specific visual knowledge on the vision side. Relying solely on language explanations is challenging for the networks to understand pathology features with limited training samples. In contrast, PEMP's learning process operates at patch and slide levels, covers visual and textual prompt design, and introduces task-specific image examples and descriptions separately as prior knowledge.

\section{FSWC Baseline Framework}
\subsection{Problem Formulation}
We provide a brief introduction to the problem definition of FSWC and the pre-trained VLM used in our study. Given a dataset $W=\left\{W_1, W_2, \ldots, W_N\right\}$ containing $N$ WSIs, each WSI $W_i$ is divided into non-overlapping small image patches $\left\{p_{i, j}, j=1,2, \ldots, n_i\right\}$, where $n_i$ represents the number of patches obtained from $W_i$. It is important to note that the number of patches obtained from each slide may vary. All patches in $W_i$ collectively form a bag, and each patch serves as an instance of this bag. Each bag is assigned a label $Y_i \in\{0,1, \ldots, U\}$, while the label of each instance is unknown, and U is the number of classes. FSWC is highly challenging as it allows training with only a limited number of slides (with labels). In the FSWC task, ``shot'' refers to the number of WSIs, e.g., $K$ pairs of WSIs and labels for training in a $K$-shot setting, while the trained model is evaluated on a completely independent test set. Typically, $K$ can take values such as 2, 4, 8, 16, or 32. 

\subsection{Attention Aggregation for WSI Classification}
\label{sec32}
The attention-based pooling strategy \cite{8} is a classical MIL aggregation method used for WSI classification. Initially, an encoder $\phi_{img}$ is used to extract features ${f}_{i, j}$ for all patches $\left\{p_{i, j}, j=1,2, \ldots n_i\right\}$ in slide $W_i$, and then aggregate these patch features using an attention-based trainable aggregation function $A(\cdot)$ to obtain the bag feature ${F}_{{i}}$:
\begin{equation}
{F}_{{i}}=A\left(\phi_{img}\left(p_{i, 1}\right), \phi_{img}\left(p_{i, 2}\right), \ldots\right)=\sum_{j=1}^{n_i} a_{i, j} f_{i, j},
  \label{eq1}
\end{equation}
\begin{equation}
a_{i, j}=\frac{\exp \left\{w^{\top} \tanh \left(V f_{i, j}^{\top}\right)\right\}}{\sum_{j=1}^{n_i} \exp \left\{w^{\top} \tanh \left(V f_{i, j}^{\top}\right)\right\}},
  \label{eq2}
\end{equation}
where $a_{i,j}$ is the attention score predicted by the self-attention network, which is parameterized by $w$ and $V$. The weights $a_{i,j}$ reflect the contribution of each patch in making the slide-level prediction. 
The patch-level encoder 
$\phi_{img}$ is from the pre-trained CLIP~\cite{53} and remains fixed during the training process.

\subsection{VLM backbone and Few-shot Prompt Learning}
We utilize CLIP~\cite{53} as the backbone of the proposed framework and also the pre-trained VLM for feature extraction, which is trained on a massive dataset of 400 million image-text pairs, including medical data, facilitating efficient knowledge transfer of diverse visual concepts to downstream tasks.

Prompt learning enables the efficient adaptation of a fixed pre-trained CLIP in few-shot downstream classification tasks by introducing task-specific prompts into the model input. These prompts, whether fixed or learnable, can be applied to the text encoder, visual encoder, or both simultaneously. In the few-shot setting, the training could also involve aligning visual and text features through alignment contrastive learning (AC-Loss) in the specific domain. Formally, we denote the visual encoder as $\phi_{img}$ and the text encoder as $\phi_{text}$. For a downstream classification task with $U$ classes, a commonly used fixed hand-crafted prompt, denoted as [Class Prompt], guides classification on the text side. Specifically, assuming the name of the $i$-th class is ``[class-name],'' the corresponding hand-crafted prompt is typically designed as ``a photo of a [class-name].'' These [Class Prompts] are pre-encoded into a vector ${C}=\left\{{c}_i\right\}_{i=1}^U$.

To further enhance generalization to downstream tasks, it is also popular to explore learning a set of continuous context vectors as learnable prompts, denoted as $\mathbb{V}=\left\{{v}_1, {v}_2, \ldots, {v}_M\right\}$. These learnable context vectors $\mathbb{V}$, concatenated with the vectors from [Class Prompt] $C$, form the complete text prompt ${t}_i=\left\{{v}_1, {v}_2, \ldots, {v}_M, {c}_i\right\}$. This prompt is then fed into the text encoder $\phi_{text}$ to obtain the final textual features for each class, $T_i=\phi_{text}\left({t}_i\right)$. On the visual side, given an image $W_i$ and its label $y$, its slide features ${F_i}$ is computed via~\cref{eq1}. Similar to the text side, learnable prompts could also be appended, which will be inputted into the visual encoder $\phi_{img}$. 

With a limited number of image-text pairs, the optimization involves minimizing the negative log-likelihood between the visual slide feature ${F_i}$ and its class text feature ${T}_y$, updating all learnable parameters. During testing, the extracted visual features ${F_i}$ are used to retrieve the class text features ${T}_y$ to obtain the probabilities for each class.
\begin{equation}
p(y \mid {F_i})=\frac{\exp \left(\operatorname{sim}\left({F_i}, {T}_y\right) / \tau\right)}{\sum_{i=1}^U \exp \left(\operatorname{sim}\left({F_i}, {T}_i\right) / \tau\right)},
  \label{eq3}
\end{equation}
where, $y$ represents the ground truth class, $U$ is the total number of classes, $\operatorname{sim}(\cdot)$ denotes cosine similarity, and $\tau$ is a learnable temperature parameter. It is crucial to note that during the training process, the parameters of CLIP's image encoder and text encoder remain fixed and unchanged.

\begin{figure*}[t]
  \centering
   \includegraphics[width=\linewidth]{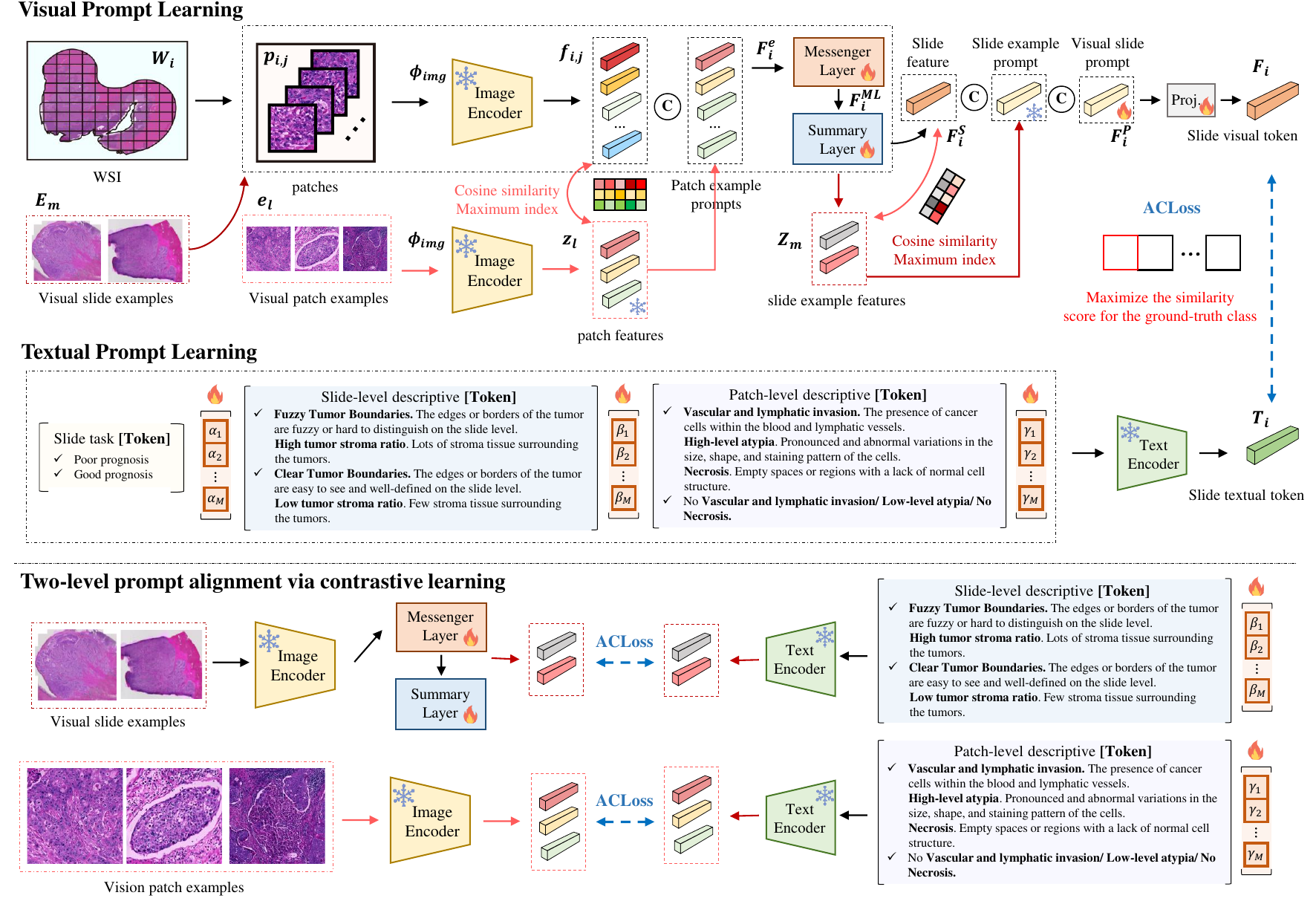}
   \caption{Overview of PEMP, where flames represent optimized parameters, and snowflakes indicate frozen parameters during training.}
   \label{figure2}
\end{figure*}

\section{Knowledge-enhanced Prompt Learning}
\subsection{Overview}
Fig.~\ref{figure2} illustrates the primary workflow of our proposed PEMP, which comprises three learning processes: \textit{Visual prompt learning}, \textit{Textual prompt learning}, and \textit{Two-level prompt alignment via contrastive learning}. In visual prompt learning, we address three key tasks. Firstly, we integrate prior visual pathology knowledge into the prompts at both patch-level and slide-level. This guides the model, particularly with limited training samples, to \textit{understand the prompt better and produce more meaningful activations for task-specific key pathology patterns}. 
Additionally, we also include the learnable prompts on the slide level, triggering the model's precise adaptation to specific tasks. Thirdly, we introduce a lightweight Messenger (self-attention) Layer and Summary (attention-pooling) Layer. The former efficiently communicates and models features of different patches within the same slide, while the latter aggregates patch features into slide features using an attention mechanism.

In a similar initiative, textual prompt learning concentrates on modeling language descriptions related to pathology tasks, also incorporating patch-level and slide-level priors. 
Pathologists are employed to provide straightforward language descriptions in the overall task, task-related slide-level pathology, and task-related patch-level pathology, avoiding obscure professional terms. 


Recognizing the challenges faced by the VLM in understanding key visual features and complex pathology terms in visual examples, we establish a close alignment between textual prompt learning (with integrated pathology language descriptions)  and visual prompt learning (with injected visual examples)  at both slide and patch levels. This alignment enhances the organization and cohesion of knowledge-enhanced prompts on both visual and textual sides.



\begin{figure}[t]
  \centering
  \begin{subfigure}[t]{0.45\textwidth}
    \centering
    \includegraphics[height=2.2cm]{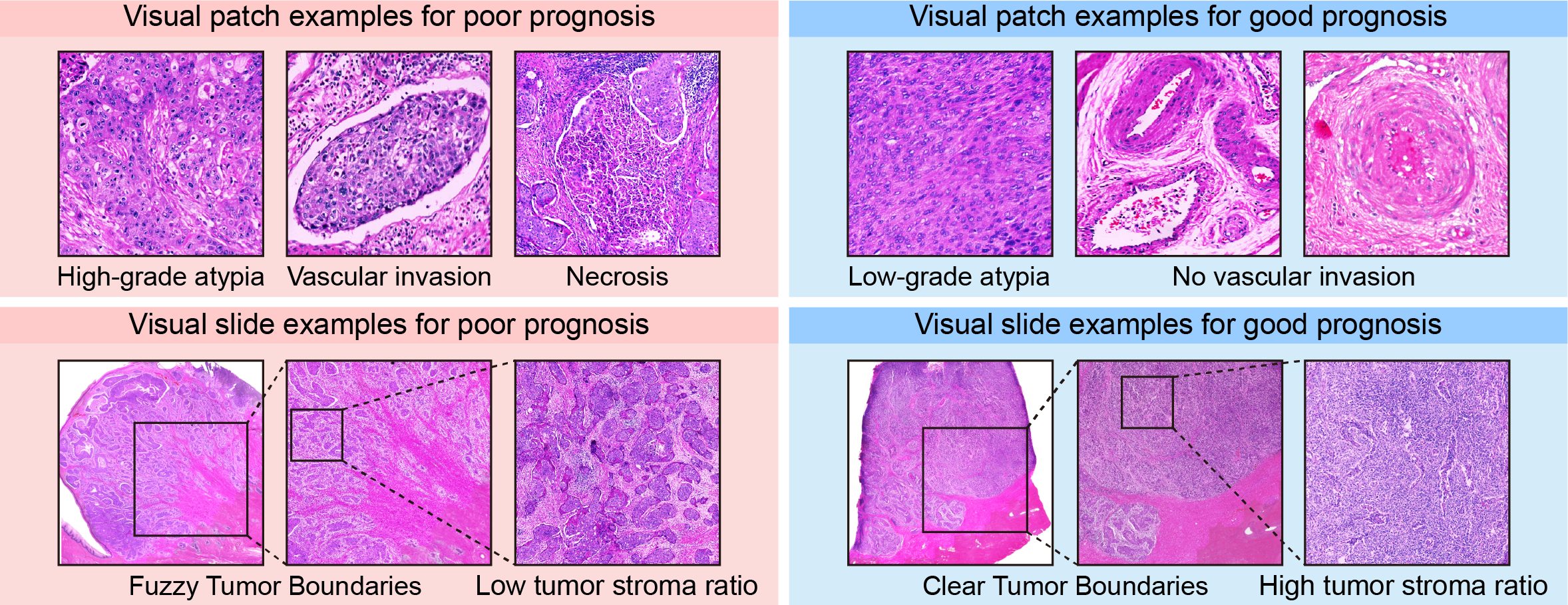}
    \caption{Examples for predicting the prognosis of early cervical cancer patients from H\&E pathology slides of the primary lesion, derived from authoritative textbooks and pathology experts. Top: visual \textit{\textbf{patch}} examples. Bottom: visual \textit{\textbf{slide}} examples.}
    \label{figure3a}
  \end{subfigure}
  \hfill
  \begin{subfigure}[t]{0.45\textwidth}
    \centering
    \includegraphics[height=2.2cm]{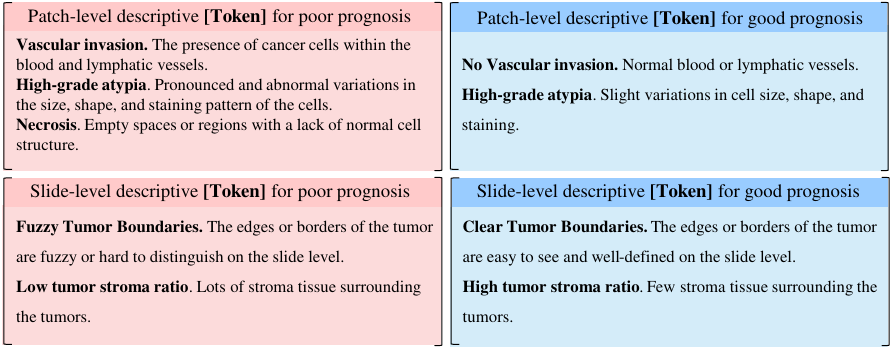}
    \caption{Top: Construction of \textbf{Patch-level} descriptive [Token]. Bottom: Construction of \textbf{Slide-level} descriptive [Token]. Examples for the prognosis of early cervical cancer patients from H\&E pathology slides of the primary lesion, serving as task-related patch-level and slide-level pathology priors on the text side.}
    \label{figure100b}
  \end{subfigure}
  \caption{(A) and (B) examples of visual and textual token constructions related to the prognosis of early cervical cancer patients.}
  \label{fig:combined}
\end{figure}

\subsection{Visual Prompt Learning}
\textbf{Construction of visual prompts}. In the FSWC task, limited training samples present challenges in acquiring effective knowledge. Moreover, tasks like survival analysis require finding key factors both locally (\eg, patch-level: necrosis, vascular invasion) and globally (\eg, slide-level: tumor boundary, tumor stroma ratio). 
To overcome these, we out-source prior visual pathology knowledge at both patch and slide levels (shown in the top part of Fig.~\ref{figure3a}), guiding the model with task-specific pathology patterns for diagnosis. For challenging slide-level diagnostic tasks, relevant patterns manifest at both levels. Leveraging expertise from pathology professionals, we construct visual \textbf{patch and slide examples} as fixed prompts, as shown in Fig.~\ref{figure3a}. For each classification task, highly relevant visual patches and slides indicating high and low risks are retrieved from authoritative sources or pathology experts. For instance, in the prognosis of early cervical cancer, visual patch examples for poor prognosis include ``high-grade atypia'', ``vascular invasion,'' and ``necrosis,'' while slide examples encompass ``overall infiltration with fuzzy tumor boundaries and low tumor stroma ratio.'' Conversely, low-risk examples include ``low-grade dysplasia,'' ``no vascular invasion,'' and ``overall pushing-type with clear tumor boundaries and high tumor stroma ratio.'' 

In total, 6 slides and 6 patch images are used as visual prompts, including 3 typical slides per prognosis category (good and poor), each with 3 local patches in the prognosis of early cervical cancer. For the Lymph Node Metastasis Prediction Task, we used 4 slides and 4 patches, with two representative slides per category (metastasis and non-metastasis) and two local images per slide. For Round Cell Subtype Diagnosis, 10 slides and 10 patches were used, with two slides per category, each with two local images. Selection of these visual examples was guided by clinical experts, with each risk factor represented by one slide and its local patch.

\noindent \textbf{Learning pipeline}. As illustrated in Fig.~\ref{figure2}, we assuming ${W}_i$ represents the training slide containing ${n}_i$ patches named ${p}_{i,j}$ and utilize the fixed CLIP image encoder $\phi_{img}$ to extract patch features $f_{i, j}={\phi}_{i m g}\left(p_{i, j}\right) \in \mathbb{R}^{1 \times \upsilon}$, where $\upsilon$ represents the dimension of the features. Subsequently, the constructed visual patch examples ${e}_l$ are also transformed into patch features $z_l=\phi_{img}\left(e_l\right)$. ${z}_l$ remains unchanged throughout the training process and is combined with the current slide's patch features as a fixed visual prompt. Specifically, we match the cosine similarity between ${f}_{i,j}$ and ${z}_l$, concatenate the most matched example patch feature with the patch feature as ${f}_{{i}, {j}}^{{e}}$, and ultimately concatenate all new patch features for the current slide ${F}_{{i}}^{{e}}=\left\{{f}_{{i}, {j}}^{{e}}\right\}_{j=1}^{n_i} \in \mathbb{R}^{n_i \times \upsilon \times 2}$. To model all patch features within a slide effectively, we introduce a lightweight self-attention layer named Messenger Layer (ML). Taking $F_i^e$ as input, it outputs $F_i^{ML}$:
\begin{equation}
F_{{i}}^{{M} {L}}=\operatorname{softmax}\left(\frac{{Q} {K}^{\top}}{\sqrt{d_w}}\right) {V},
  \label{eq4}
\end{equation}
where $Q$, $K$, and $V$ are Query, Key, and Value vectors obtained through linear mapping of $F_i^e$, and $d_w$ is the feature dimension. Next, we construct a lightweight Attention Aggregation Layer to aggregate all patch features $F_i^{ML}$ within a slide into a slide feature $F_i^S$. The attention weights are learnable, and the detailed structure is shown in Section \ref{sec32}. Following this, the constructed visual slide examples $E_m$ undergo the aforementioned process like $W_i$ (i.e., patch extraction, patch example matching, Messenger Layer, and Summary Layer) to obtain their slide features $Z_m$. These slide example features are also integrated as prompts into slide-level training. Similar to the patch-level matching, we match the cosine similarity between $F_i^S$ and $Z_m$ and concatenate the most matched slide example feature with the slide feature. Additionally, we introduce learnable vector $F_i^P$ as a slide-level learnable prompt, concatenate it together, and input them into a one-layer fully-connected layer named ``projector'' to adjust the feature dimension, yielding the final slide visual features ${F}_{{i}} \in \mathbb{R}^{1 \times \upsilon}$.

\subsection{Textual Prompt Learning}
\textbf{Construction of textual prompts}. Textual prompt learning primarily focuses on modeling language descriptions related to pathology tasks, patch-level, and slide-level pathology priors on the text side. This knowledge is integrated into the training process through fixed descriptive prompts and learnable prompts (shown in the middle of Fig.~\ref{figure2}). The construction of textual prompts could be divided into mainly three parts. 
The following detailed explanation uses the task ``predicting the prognosis of early cervical cancer patients from H\&E pathology slides of the primary lesion'' as an example, also illustrated in Fig.~\ref{figure100b}.


First, the ``\textbf{Slide task [Token]}'' primarily serves to prompt the main category of the current task, such as ``A Whole Slide Image of cervical cancer from primary tumors with a [poor/good] prognosis.'' Furthermore, we concatenate learnable prompts ($\left([\alpha]_1[\alpha]_2 \ldots[\alpha]_M\right)$) with the ``Slide task [Token]'' to adaptively adjust and complement during few-shot training.

Then, the ``\textbf{Slide-level descriptive [Token]}'' is a straightforward language description corresponding to the aforementioned visual slide examples. These descriptions are provided by experienced pathology experts, avoiding obscure professional terminology. For instance, the description corresponding to the ``Poor prognosis'' category is: ``Fuzzy Tumor Boundaries. The edges or borders of the tumor are fuzzy or hard to distinguish on the slide level. Low tumor stroma ratio. Few stroma tissues surrounding the tumors.'' In contrast, the description corresponding to the ``Good prognosis'' category is: ``Clear Tumor Boundaries. The edges or borders of the tumor are easy to see and well-defined on the slide level. High tumor stroma ratio. Lots of stroma tissue surrounding the tumors.'' Similarly, we also append learnable prompts ($\left([\beta]_1[\beta]_2 \ldots[\beta]_M\right)$).

Finally, the ``\textbf{Patch-level descriptive [Token]}'' provides the language description corresponding to the aforementioned visual patch examples. For example, the description corresponding to the ``Poor prognosis'' category is: ``Vascular and lymphatic invasion: The presence of cancer cells within the blood and lymphatic vessels. High-level atypia: Pronounced and abnormal variations in the size, shape, and staining pattern of the cells. Necrosis: Empty spaces or regions with a lack of normal cell structure.'' Similarly, immediately following this, we also include learnable prompts ($\left([\gamma]_1[\gamma]_2 \ldots[\gamma]_M\right)$).

It is crucial to note that the second and third parts correspond closely to the same number of visual slide and patch examples in visual prompt learning. We input each part's visual descriptions and learnable vectors together into the CLIP Text Encoder, obtaining the final slide textual features ${T}_{{i}} \in \mathbb{R}^{1 \times \upsilon}$.

\subsection{Alignment of Visual and Textual Prompts }
Alignment contrastive learning (AC-Loss) is originally used to align corresponding visual representation and textual representation. 
Nonetheless, the overall loss function in this paper consists of three parts of AC-Loss, expressed as $\mathcal{L}_{\text {total }}=\mathcal{L}_t+\lambda_1 \mathcal{L}_s+\lambda_2 \mathcal{L}_p$. \textit{Conceptually}, $\mathcal{L}_t$ aligns the overall slide visual features $F_i$ with the slide textual features $T_i$ corresponding to the task category, thereby completing slide-level classification. $\mathcal{L}_s$ aligns the features of slide-level visual examples in visual prompt learning with the slide-level pathological language descriptions in textual prompt learning. $\mathcal{L}_p$ aligns the features of patch-level visual examples with patch-level pathology language descriptions (shown in the bottom of Fig.~\ref{figure2}).

Mathematically, the forms of these three parts of AC-Loss are identical. Here, we take $\mathcal{L}_t$ as an example. AC-Loss optimizes the learnable parameters by minimizing the negative log-likelihood between the image features $F_i$ and the ground truth class text features $T_y$. Its basic form is as follows:
\begin{equation}
\mathcal{L}=-\sum_{F_i} \log \frac{\exp \left(\operatorname{sim}\left({F_i}, {T}_y\right) / \tau\right)}{\sum_{i=1}^U \exp \left(\operatorname{sim}\left({F_i}, {T}_i\right) / \tau\right)},
  \label{eq5}
\end{equation}
where $y$ is the ground truth category, $U$ is the total number of categories, $\operatorname{sim}(\cdot)$ denotes the cosine similarity, and\ $\tau$ is a learnable temperature parameter. 

During testing, we input the unseen WSI and obtain its slide visual feature in the same fashion as training. Subsequently, we match it separately with the textual features for ``poor prognosis'' and ``good prognosis'' and use softmax to obtain the predicted risks for the two categories.

\section{Experiment and Result}
\noindent\textbf{Datasets and Evaluation Metrics.}
This paper addresses clinical tasks in H\&E pathology diagnosis, specifically focusing on data label scarcity, the complexity of direct diagnosis by pathologists, and limited sample sizes for rare diseases. 
This study encompasses three clinical tasks for five tumor types across five datasets: (1) Predicting patient survival prognosis and lymph node metastasis from H\&E slides of early-stage cervical cancer primary lesions using an in-house dataset (For testing, 80 cases and 74 cases, individually) and a public TCGA-CESC dataset (For testing, 80 cases and 70 cases, individually). Note that the prompts utilized in the public TCGA datasets are the same as those used in the corresponding in-house datasets for each task. We ensure no overlap between our training data and visual prompts (obtained from external authoritative sources or pathology experts) and testing data, across both in-house and TCGA datasets. Furthermore, the lymph node metastasis detection tasks from primary lesions assessed in this study present a significantly greater challenge compared to those from sentinel lymph nodes, as utilized in prior research, with samples and labels being notably rarer. (2) Classifying round cell tumors, including neuroendocrine tumors, malignant melanoma, lymphohematopoietic tumors, and soft tissue tumors, into subtypes using an in-house dataset (For testing, 112 cases). More details are in the Supplementary Materials.
For the prognosis prediction task in cervical cancer, the concordance index (C-index) is used, while for other tasks, the Area Under the Curve (AUC) is employed. Higher values for both metrics indicate improved performance.

\noindent\textbf{Compared Methods.}
Currently, few methods directly address FSWC tasks through prompt learning, with the SOTA being represented by the recently proposed TOP method \cite{83} in the pathology domain. In the natural domain, prompt learning work based on VLMs is limited to single image patches and does not readily apply to slide-level classification. Following the TOP approach, we adopt advanced prompt learning methods from the natural domain for the MIL problem. These methods include: \textit{(1) Linear-Probe} \cite{53}, \textit{(2) VPT} \cite{91}, \textit{(3) CoOp} \cite{56}, and \textit{(4) KgCoOp} \cite{86}. 
The Messenger Layer and Summary Layer, as described in this paper, aggregate patch features into a slide feature. All compared methods use the same Messenger \& Summary layers for fair comparisons. For Linear-Probe, a linear layer performs slide-level classification on the aggregated feature. VPT incorporates a learnable prompt on the vision side for slide-level classification. CoOp utilizes CLIP's prompt learning by adding the Slide Task [Token] and learnable textual prompts trained through AC-Loss. KgCoOp imposes constraints on learnable textual prompts. Few-shot experiments are conducted for each category with 2, 4, 8, 16, and 32 slides (with labels). Results are reported as an average over five experimental runs. Standard deviation of our main results are provided in the supplementary materials.

\begin{table}[t!]
  \centering
    \caption{Comparison for Survival Prognosis Prediction Task.}
      \setlength{\tabcolsep}{0.2cm}
      \resizebox{0.8\textwidth}{!}{
\begin{tabular}{c|ccccc|ccccc}
\hline
Dataset              & \multicolumn{5}{c|}{In-house Dataset}                                              & \multicolumn{5}{c}{Public TCGA-CESC   Dataset}                                     \\ \hline
Method               & 32-shot        & 16-shot        & 8-shot         & 4-shot         & 2-shot         & 32-shot        & 16-shot        & 8-shot         & 4-shot         & 2-shot         \\ \hline
LinearProbe          & 0.620          & 0.562          & 0.543          & 0.501          & 0.458          & 0.584          & 0.577          & 0.543          & 0.511          & 0.485          \\
VPT(ECCV'22)         & 0.626          & 0.569          & 0.545          & 0.501          & 0.464          & 0.589          & 0.573          & 0.550          & 0.518          & 0.482          \\
CoOp(IJCV'22)        & 0.641          & 0.594          & 0.561          & 0.517          & 0.490          & 0.604          & 0.585          & 0.558          & 0.523          & 0.487          \\
KgCoOp(CVPR'23)      & 0.635          & 0.587          & 0.552          & 0.513          & 0.487          & 0.597          & 0.584          & 0.550          & 0.519          & 0.486          \\
TOP(NeurIPS'23)      & 0.652          & 0.608          & 0.574          & 0.539          & 0.508          & 0.611          & 0.597          & 0.566          & 0.536          & 0.518          \\ \hline
\textbf{PEMP (ours)} & \textbf{0.667} & \textbf{0.637} & \textbf{0.614} & \textbf{0.587} & \textbf{0.562} & \textbf{0.637} & \textbf{0.624} & \textbf{0.602} & \textbf{0.577} & \textbf{0.551} \\ \hline
\end{tabular}}
  \label{table1}
\end{table}

\begin{table}[t!]
  \centering
    \caption{Comparison for Lymph Node Metastasis Prediction Task.}
          \setlength{\tabcolsep}{0.2cm}
      \resizebox{0.8\textwidth}{!}{
\begin{tabular}{c|ccccc|ccccc}
\hline
Dataset              & \multicolumn{5}{c|}{In-house Dataset}                                              & \multicolumn{5}{c}{Public TCGA-CESC   Dataset}                                     \\ \hline
Method               & 32-shot        & 16-shot        & 8-shot         & 4-shot         & 2-shot         & 32-shot        & 16-shot        & 8-shot         & 4-shot         & 2-shot         \\ \hline
LinearProbe          & 0.781          & 0.735          & 0.728          & 0.716          & 0.687          & 0.763          & 0.719          & 0.696          & 0.671          & 0.655          \\
VPT(ECCV'22)         & 0.785          & 0.742          & 0.728          & 0.719          & 0.689          & 0.769          & 0.724          & 0.712          & 0.678          & 0.655          \\
CoOp(IJCV'22)        & 0.792          & 0.757          & 0.735          & 0.727          & 0.693          & 0.786          & 0.741          & 0.723          & 0.691          & 0.659          \\
KgCoOp(CVPR'23)      & 0.789          & 0.753          & 0.731          & 0.724          & 0.693          & 0.775          & 0.733          & 0.710          & 0.687          & 0.652          \\
TOP(NeurIPS'23)      & 0.825          & 0.819          & 0.801          & 0.787          & 0.762          & 0.799          & 0.761          & 0.744          & 0.708          & 0.679          \\ \hline
\textbf{PEMP (ours)} & \textbf{0.849} & \textbf{0.838} & \textbf{0.824} & \textbf{0.801} & \textbf{0.783} & \textbf{0.818} & \textbf{0.795} & \textbf{0.760} & \textbf{0.726} & \textbf{0.704} \\ \hline
\end{tabular}}
  \label{table2}
\end{table}

\begin{table}[t!]
  \centering
    \caption{Comparison for Round Cell Subtype Diagnosis Task.}
      \setlength{\tabcolsep}{0.2cm}
\resizebox{0.6\textwidth}{!}{
\begin{tabular}{lccccc}
\cline{1-6}
Method               & 32-shot        & 16-shot        & 8-shot         & 4-shot         & 2-shot       \\ \cline{1-6}
LinearProbe          & 0.654          & 0.606          & 0.549          & 0.517          & 0.483        \\
VPT(ECCV'22)         & 0.659          & 0.613          & 0.556          & 0.515          & 0.511        \\
CoOp(IJCV'22)        & 0.667          & 0.627          & 0.568          & 0.553          & 0.549        \\
KgCoOp(CVPR'23)      & 0.661          & 0.619          & 0.547          & 0.535          & 0.534        \\
TOP(NeurIPS'23)      & 0.682          & 0.652          & 0.633          & 0.584          & 0.560        \\ \cline{1-6}
\textbf{PEMP (ours)} & \textbf{0.751} & \textbf{0.718} & \textbf{0.685} & \textbf{0.643} & \textbf{0.625} \\ \cline{1-6}
\end{tabular}}
  \label{table3}
\end{table}

\subsection{Results}
\noindent\textbf{Survival prognosis prediction.} Predicting survival prognosis, especially directly from H\&E slides with a limited number of samples, poses a significant challenge. Table \ref{table1} showcases the performance of all methods in the in-house dataset and the public TCGA-CESC dataset, with PEMP consistently demonstrating superior performance across all settings with few samples, outperforming all comparative methods significantly. In scenarios with limited samples, LinearProbe and VPT, relying solely on features extracted by the CLIP Image Encoder on the visual side without incorporating language-based prior knowledge, exhibit suboptimal performance. CoOp and KgCoOp, which combine visual and language aspects, struggle to comprehend complex medical terms related to prognosis effectively, limiting their performance. TOP enhances language descriptions at both slide and patch levels but provides explanations solely on the language side, making it challenging to establish a direct relationship between crucial visual features and language descriptions. In contrast, PEMP introduces typical patch and slide examples related to prognosis on the visual side, along with straightforward language descriptions on the language side. Through AC-Loss, the model efficiently completes the task under the guidance of pathology prior knowledge. Additionally, PEMP excels when the sample size is very small (\eg, 2-shot, 4-shot), showcasing its effective utilization of visual and language prior knowledge related to prognosis in H\&E pathology slides for diagnosis.

\noindent\textbf{Lymph node metastasis prediction.} Table \ref{table2} presents the performance of all methods in the in-house and TCGA-CESC datasets. Across all settings with limited samples, PEMP consistently achieves the best performance and significantly outperforms all comparative methods. This indicates that PEMP is also capable of efficiently understanding and leveraging pathological prior knowledge related to metastasis in primary lesion H\&E pathology slides for diagnosis. The results of both tasks highlight the immense potential of PEMP in clinical tasks characterized by challenges in label acquisition and diagnostic complexity.

\noindent\textbf{Round cell subtype diagnosis.} The classification diagnosis for rare diseases in the round cell subtype task presents an increasingly challenging yet vital clinical issue. Table \ref{table3} illustrates the performance of all methods. Notably, with a limited number of shots, most methods face challenges in achieving satisfactory diagnostic results. However, leveraging visual and pathological expertise from pathology experts, PEMP achieves significantly better results for all the settings (6.2\% higher AUC on average). This underscores the substantial potential of PEMP in the diagnosis of rare diseases.

\begin{table}[t!]
  \centering
    \caption{Ablation studies of the modules.}
      \setlength{\tabcolsep}{0.2cm}
\resizebox{0.6\textwidth}{!}{
\begin{tabular}{lccccc}
\cline{1-6}
Method               & 32-shot        & 16-shot        & 8-shot         & 4-shot         & 2-shot       \\ \cline{1-6}
PEMP w/o v\&t em.          & 0.641          & 0.594          & 0.561          & 0.517          & 0.490        \\
PEMP vision only         & 0.649          & 0.607          & 0.568          & 0.543          & 0.527        \\
PEMP w/o vision em.        & 0.655          & 0.613          & 0.577          & 0.540          & 0.511        \\
PEMP w/o text em.      & 0.658          & 0.619          & 0.575          & 0.557          & 0.533        \\
PEMP w/o Summary Layer    & 0.632          & 0.587          & 0.558          & 0.511          & 0.487          \\
PEMP w/o Messenger Layer   & 0.664 & 0.629          & 0.594          & 0.575          & 0.554          \\
PEMP w/o Slide-level Prompts & 0.656 & 0.620           & 0.581          & 0.552          & 0.525          \\
PEMP w/o AC-Loss      & 0.660          & 0.626          & 0.589          & 0.568          & 0.549        \\ \cline{1-6}
\textbf{PEMP (ours)} & \textbf{0.667} & \textbf{0.637} & \textbf{0.614} & \textbf{0.587} & \textbf{0.562} \\ \cline{1-6}
\end{tabular}}
  \label{table4}
\end{table}

\noindent\textbf{Ablation Study.} As shown in Table~\ref{table4}, we designed eight variants and evaluated the roles of each component in PEMP in the prognosis task. (1) PEMP w/o v\&t em. represents the removal of all vision and text examples, reducing PEMP to CoOp; (2) PEMP vision only denotes the exclusion of the text end, training solely with vision examples as guidance for prior knowledge; (3) PEMP w/o vision em. indicates the removal of all vision examples; (4) PEMP w/o text em. signifies the exclusion of all text examples; (5) PEMP w/o Summary Layer denotes the replacement from Summary Layer to simple average pooling. Please note that the Summary Layer is essentially an Attention Pooling layer, recognized as a crucial component in MIL problems; (6) PEMP w/o Messenger Layer denotes the removal of the Messenger Layer; (7) PEMP w/o Slide-level Prompt denotes the removal of all slide-level prompts; (8) PEMP w/o AC-Loss denotes the removal of AC-Loss for vision and text examples while retaining the AC-Loss for the visual and text features of the slide task token. 

\begin{figure*}[t!]
  \centering
   \includegraphics[width=0.7\linewidth]{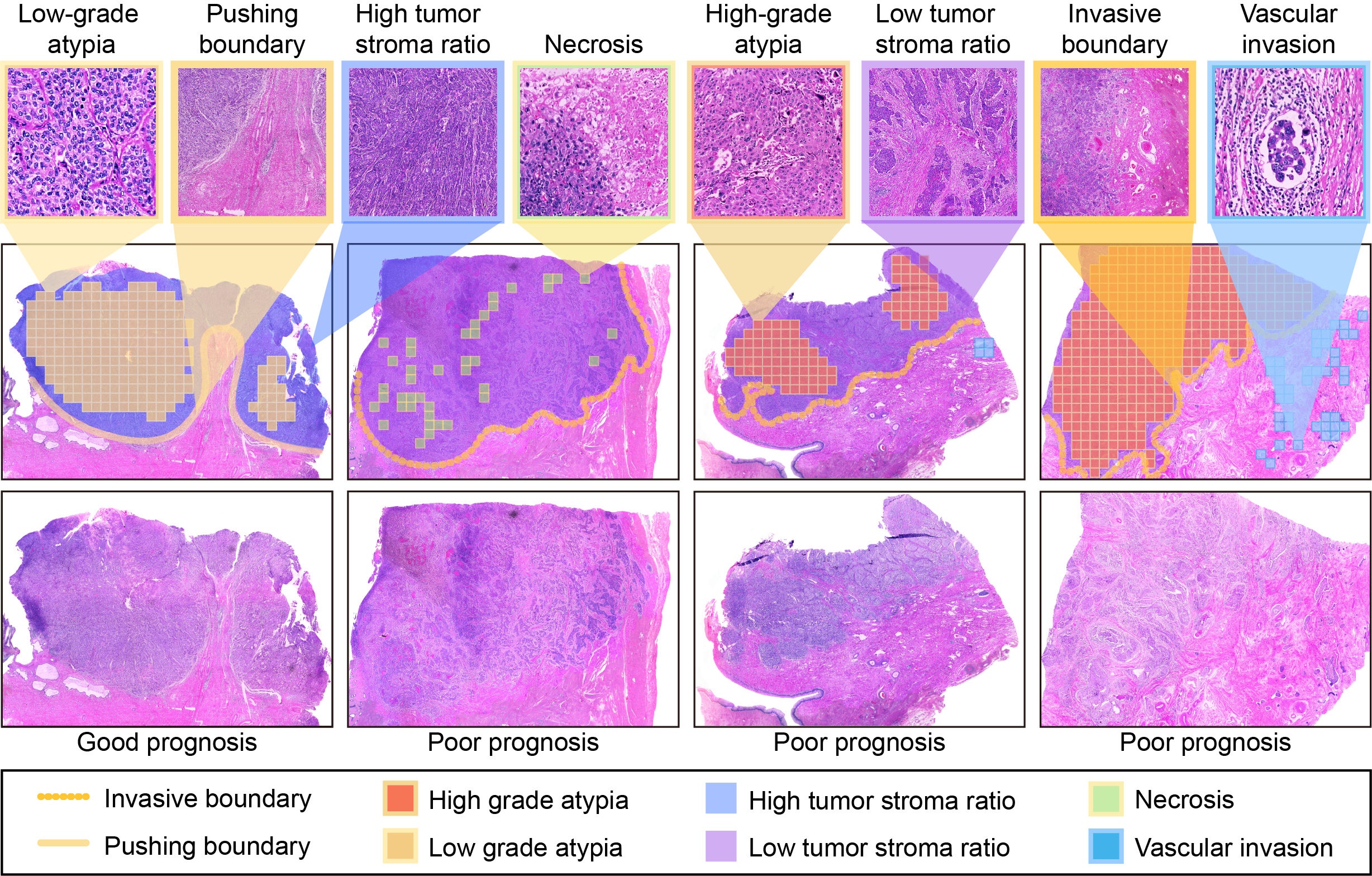}
   \caption{Visualization of key pathology patterns indicating both good and poor prognosis retrieved by PEMP from the test set.}
   \label{figure4}
\end{figure*}

\noindent\textbf{Visualization and Interpretability.}
To illustrate the robust performance of PEMP, we briefly visualize typical cases where visual patch examples and slide examples exhibit the highest similarity on the prognosis task's test dataset (\cref{figure4}). PEMP demonstrates accurate learning of pathology prior knowledge, effectively addressing the challenge of predicting survival outcomes with limited training WSIs. More importantly, PEMP also shows high interpretability. PEMP efficiently integrates the high-risk and low-risk pathology image patterns and textual descriptions that pathologists have summarized over an extended period. PEMP achieves excellent performance under the guidance and facilitation of this prior knowledge.


\section{Conclusion}
This paper presents PEMP, a pathology-knowledge enhanced multi-instance prompt learning framework, excelling in challenging tasks such as prognosis prediction with difficult label acquisition and rare tumor classification in Few-shot Weakly Supervised WSI Classification. PEMP integrates vision and text prior knowledge into prompts at both patch and slide levels, using a combination of static and learnable prompts during training to guide the effective diagnosis of key pathology patterns. 
Alignment-wise contrastive losses ensure feature-level alignment between knowledge-enhanced visual and textual prompts at both patch and slide levels.
In specific clinical settings, collecting a large number of WSIs may be challenging due to patient privacy issues, the prevalence of rare diseases, new treatment protocols, or small patient groups. PEMP has shown substantial benefits across five datasets, including prognosis and rare diseases, demonstrating significant diagnostic benefits in fields with limited medical resources.
PEMP also demonstrates high interpretability in medical diagnoses, indicating significant potential for future clinical applications. 
Moreover, such prompt-description-based model adaptation could be widely adopted as a general learning paradigm for specialized domain applications.

\section*{Acknowledgements}
This work is funded by the National Key R\&D Program of China (2022ZD0160700) and Shanghai AI Laboratory.

%
%
\bibliographystyle{splncs04}
\bibliography{main1}

\begin{thebibliography}{10}
\providecommand{\url}[1]{\texttt{#1}}
\providecommand{\urlprefix}{URL }
\providecommand{\doi}[1]{https://doi.org/#1}

\bibitem{85}
Alayrac, J.B., Donahue, J., Luc, P., Miech, A., Barr, I., Hasson, Y., Lenc, K., Mensch, A., Millican, K., Reynolds, M., et~al.: Flamingo: a visual language model for few-shot learning. Advances in Neural Information Processing Systems (NeurIPS)  \textbf{35},  23716--23736 (2022)

\bibitem{1}
Campanella, G., Hanna, M.G., Geneslaw, L., Miraflor, A., Werneck Krauss~Silva, V., Busam, K.J., Brogi, E., Reuter, V.E., Klimstra, D.S., Fuchs, T.J.: Clinical-grade computational pathology using weakly supervised deep learning on whole slide images. Nature Medicine  \textbf{25}(8),  1301--1309 (2019)

\bibitem{92}
Chan, T.H., Cendra, F.J., Ma, L., Yin, G., Yu, L.: Histopathology whole slide image analysis with heterogeneous graph representation learning. In: Proceedings of the IEEE/CVF Conference on Computer Vision and Pattern Recognition (CVPR). pp. 15661--15670 (2023)

\bibitem{95}
Chen, R.J., Chen, C., Li, Y., Chen, T.Y., Trister, A.D., Krishnan, R.G., Mahmood, F.: Scaling vision transformers to gigapixel images via hierarchical self-supervised learning. In: Proceedings of the IEEE/CVF Conference on Computer Vision and Pattern Recognition (CVPR). pp. 16144--16155 (2022)

\bibitem{34}
Chen, R.J., Lu, M.Y., Wang, J., Williamson, D.F., Rodig, S.J., Lindeman, N.I., Mahmood, F.: Pathomic fusion: an integrated framework for fusing histopathology and genomic features for cancer diagnosis and prognosis. IEEE Transactions on Medical Imaging  \textbf{41}(4),  757--770 (2020)

\bibitem{30}
Chen, R.J., Lu, M.Y., Weng, W.H., Chen, T.Y., Williamson, D.F., Manz, T., Shady, M., Mahmood, F.: Multimodal co-attention transformer for survival prediction in gigapixel whole slide images. In: Proceedings of the IEEE/CVF International Conference on Computer Vision (ICCV). pp. 4015--4025 (2021)

\bibitem{88}
Chen, W., Si, C., Zhang, Z., Wang, L., Wang, Z., Tan, T.: Semantic prompt for few-shot image recognition. In: Proceedings of the IEEE/CVF Conference on Computer Vision and Pattern Recognition (CVPR). pp. 23581--23591 (2023)

\bibitem{94}
Chen, Y.C., Lu, C.S.: Rankmix: Data augmentation for weakly supervised learning of classifying whole slide images with diverse sizes and imbalanced categories. In: Proceedings of the IEEE/CVF Conference on Computer Vision and Pattern Recognition (CVPR). pp. 23936--23945 (2023)

\bibitem{27}
Cheplygina, V., de~Bruijne, M., Pluim, J.P.: Not-so-supervised: a survey of semi-supervised, multi-instance, and transfer learning in medical image analysis. Medical Image Analysis  \textbf{54},  280--296 (2019)

\bibitem{2}
Chikontwe, P., Kim, M., Nam, S.J., Go, H., Park, S.H.: Multiple instance learning with center embeddings for histopathology classification. In: Medical Image Computing and Computer Assisted Intervention (MICCAI). pp. 519--528. Springer (2020)

\bibitem{89}
Gu, J., Han, Z., Chen, S., Beirami, A., He, B., Zhang, G., Liao, R., Qin, Y., Tresp, V., Torr, P.: A systematic survey of prompt engineering on vision-language foundation models. arXiv preprint arXiv:2307.12980  (2023)

\bibitem{9}
Hashimoto, N., Fukushima, D., Koga, R., Takagi, Y., Ko, K., Kohno, K., Nakaguro, M., Nakamura, S., Hontani, H., Takeuchi, I.: Multi-scale domain-adversarial multiple-instance cnn for cancer subtype classification with unannotated histopathological images. In: Proceedings of the IEEE/CVF Conference on Computer Vision and Pattern Recognition (CVPR). pp. 3852--3861 (2020)

\bibitem{97}
Huang, Y., Zhao, W., Wang, S., Fu, Y., Jiang, Y., Yu, L.: Conslide: Asynchronous hierarchical interaction transformer with breakup-reorganize rehearsal for continual whole slide image analysis. In: Proceedings of the IEEE/CVF International Conference on Computer Vision (ICCV). pp. 21349--21360 (2023)

\bibitem{67}
Huang, Z., Bianchi, F., Yuksekgonul, M., Montine, T.J., Zou, J.: A visual--language foundation model for pathology image analysis using medical twitter. Nature Medicine pp. 1--10 (2023)

\bibitem{100}
Ikezogwo, W., Seyfioglu, S., Ghezloo, F., Geva, D., Sheikh~Mohammed, F., Anand, P.K., Krishna, R., Shapiro, L.: Quilt-1m: One million image-text pairs for histopathology. Advances in Neural Information Processing Systems (NeurIPS)  \textbf{36} (2024)

\bibitem{8}
Ilse, M., Tomczak, J., Welling, M.: Attention-based deep multiple instance learning. In: International Conference on Machine Learning (ICML). pp. 2127--2136. PMLR (2018)

\bibitem{91}
Jia, M., Tang, L., Chen, B.C., Cardie, C., Belongie, S., Hariharan, B., Lim, S.N.: Visual prompt tuning. In: European Conference on Computer Vision (ECCV). pp. 709--727. Springer (2022)

\bibitem{13}
Li, B., Li, Y., Eliceiri, K.W.: Dual-stream multiple instance learning network for whole slide image classification with self-supervised contrastive learning. In: Proceedings of the IEEE/CVF Conference on Computer Vision and Pattern Recognition (CVPR). pp. 14318--14328 (2021)

\bibitem{16}
Li, H., Yang, F., Zhao, Y., Xing, X., Zhang, J., Gao, M., Huang, J., Wang, L., Yao, J.: Dt-mil: deformable transformer for multi-instance learning on histopathological image. In: Medical Image Computing and Computer Assisted Intervention (MICCAI). pp. 206--216. Springer (2021)

\bibitem{93}
Li, H., Zhu, C., Zhang, Y., Sun, Y., Shui, Z., Kuang, W., Zheng, S., Yang, L.: Task-specific fine-tuning via variational information bottleneck for weakly-supervised pathology whole slide image classification. In: Proceedings of the IEEE/CVF Conference on Computer Vision and Pattern Recognition (CVPR). pp. 7454--7463 (2023)

\bibitem{84}
Li, J., Li, D., Xiong, C., Hoi, S.: Blip: Bootstrapping language-image pre-training for unified vision-language understanding and generation. In: International Conference on Machine Learning (ICML). pp. 12888--12900. PMLR (2022)

\bibitem{50}
Lin, T., Xu, H., Yang, C., Xu, Y.: Interventional multi-instance learning with deconfounded instance-level prediction. In: Proceedings of the AAAI Conference on Artificial Intelligence (AAAI). vol.~36, pp. 1601--1609 (2022)

\bibitem{51}
Lin, T., Yu, Z., Hu, H., Xu, Y., Chen, C.W.: Interventional bag multi-instance learning on whole-slide pathological images. In: Proceedings of the IEEE/CVF Conference on Computer Vision and Pattern Recognition (CVPR). pp. 19830--19839 (2023)

\bibitem{66}
Lu, M.Y., Chen, B., Zhang, A., Williamson, D.F., Chen, R.J., Ding, T., Le, L.P., Chuang, Y.S., Mahmood, F.: Visual language pretrained multiple instance zero-shot transfer for histopathology images. In: Proceedings of the IEEE/CVF Conference on Computer Vision and Pattern Recognition (CVPR). pp. 19764--19775 (2023)

\bibitem{37}
Lu, M.Y., Williamson, D.F., Chen, T.Y., Chen, R.J., Barbieri, M., Mahmood, F.: Data-efficient and weakly supervised computational pathology on whole-slide images. Nature Biomedical Engineering  \textbf{5}(6),  555--570 (2021)

\bibitem{24}
Qu, L., Liu, S., Liu, X., Wang, M., Song, Z.: Towards label-efficient automatic diagnosis and analysis: a comprehensive survey of advanced deep learning-based weakly-supervised, semi-supervised and self-supervised techniques in histopathological image analysis. Physics in Medicine \& Biology  (2022)

\bibitem{83}
Qu, L., Luo, X., Fu, K., Wang, M., Song, Z.: The rise of ai language pathologists: Exploring two-level prompt learning for few-shot weakly-supervised whole slide image classification. arXiv preprint arXiv:2305.17891  (2023)

\bibitem{3}
Qu, L., Luo, X., Liu, S., Wang, M., Song, Z.: Dgmil: Distribution guided multiple instance learning for whole slide image classification. In: Medical Image Computing and Computer Assisted Intervention (MICCAI). pp. 24--34. Springer (2022)

\bibitem{53}
Radford, A., Kim, J.W., Hallacy, C., Ramesh, A., Goh, G., Agarwal, S., Sastry, G., Askell, A., Mishkin, P., Clark, J., et~al.: Learning transferable visual models from natural language supervision. In: International Conference on Machine Learning (ICML). pp. 8748--8763. PMLR (2021)

\bibitem{26}
Rony, J., Belharbi, S., Dolz, J., Ayed, I.B., McCaffrey, L., Granger, E.: Deep weakly-supervised learning methods for classification and localization in histology images: a survey. arXiv preprint arXiv:1909.03354  (2019)

\bibitem{15}
Shao, Z., Bian, H., Chen, Y., Wang, Y., Zhang, J., Ji, X., et~al.: Transmil: Transformer based correlated multiple instance learning for whole slide image classification. Advances in Neural Information Processing Systems (NeurIPS)  \textbf{34},  2136--2147 (2021)

\bibitem{12}
Shi, X., Xing, F., Xie, Y., Zhang, Z., Cui, L., Yang, L.: Loss-based attention for deep multiple instance learning. In: Proceedings of the AAAI Conference on Artificial Intelligence (AAAI). vol.~34, pp. 5742--5749 (2020)

\bibitem{96}
Song, A.H., Jaume, G., Williamson, D.F., Lu, M.Y., Vaidya, A., Miller, T.R., Mahmood, F.: Artificial intelligence for digital and computational pathology. Nature Reviews Bioengineering pp. 1--20 (2023)

\bibitem{52}
Tu, C., Zhang, Y., Ning, Z.: Dual-curriculum contrastive multi-instance learning for cancer prognosis analysis with whole slide images. Advances in Neural Information Processing Systems (NeurIPS)  \textbf{35},  29484--29497 (2022)

\bibitem{42}
Wang, X., Xiang, J., Zhang, J., Yang, S., Yang, Z., Wang, M.H., Zhang, J., Yang, W., Huang, J., Han, X.: Scl-wc: Cross-slide contrastive learning for weakly-supervised whole-slide image classification. Advances in Neural Information Processing Systems (NeurIPS)  \textbf{35},  18009--18021 (2022)

\bibitem{87}
Wasim, S.T., Naseer, M., Khan, S., Khan, F.S., Shah, M.: Vita-clip: Video and text adaptive clip via multimodal prompting. In: Proceedings of the IEEE/CVF Conference on Computer Vision and Pattern Recognition (CVPR). pp. 23034--23044 (2023)

\bibitem{29}
Xu, G., Song, Z., Sun, Z., Ku, C., Yang, Z., Liu, C., Wang, S., Ma, J., Xu, W.: Camel: A weakly supervised learning framework for histopathology image segmentation. In: Proceedings of the IEEE/CVF International Conference on Computer Vision (ICCV). pp. 10682--10691 (2019)

\bibitem{86}
Yao, H., Zhang, R., Xu, C.: Visual-language prompt tuning with knowledge-guided context optimization. In: Proceedings of the IEEE/CVF Conference on Computer Vision and Pattern Recognition (CVPR). pp. 6757--6767 (2023)

\bibitem{11}
Yao, J., Zhu, X., Jonnagaddala, J., Hawkins, N., Huang, J.: Whole slide images based cancer survival prediction using attention guided deep multiple instance learning networks. Medical Image Analysis  \textbf{65},  101789 (2020)

\bibitem{14}
Zhang, H., Meng, Y., Zhao, Y., Qiao, Y., Yang, X., Coupland, S.E., Zheng, Y.: Dtfd-mil: Double-tier feature distillation multiple instance learning for histopathology whole slide image classification. In: Proceedings of the IEEE/CVF Conference on Computer Vision and Pattern Recognition (CVPR). pp. 18802--18812 (2022)

\bibitem{90}
Zhang, Y., Gao, J., Zhou, M., Wang, X., Qiao, Y., Zhang, S., Wang, D.: Text-guided foundation model adaptation for pathological image classification. In: International Conference on Medical Image Computing and Computer-Assisted Intervention (MICCAI). pp. 272--282. Springer (2023)

\bibitem{56}
Zhou, K., Yang, J., Loy, C.C., Liu, Z.: Learning to prompt for vision-language models. International Journal of Computer Vision  \textbf{130}(9),  2337--2348 (2022)

\bibitem{10}
Zhu, X., Yao, J., Zhu, F., Huang, J.: Wsisa: Making survival prediction from whole slide histopathological images. In: Proceedings of the IEEE/CVF Conference on Computer Vision and Pattern Recognition (CVPR). pp. 7234--7242 (2017)

\end{thebibliography}
\end{document}